\documentclass[runningheads]{llncs}
\usepackage{graphicx}
\usepackage{cite}
\usepackage{amsmath,amssymb,amsfonts}
\usepackage{algorithm}
\usepackage[noend]{algpseudocode}
\usepackage{textcomp}
\usepackage{subcaption,siunitx,booktabs,dcolumn}
\usepackage{nicematrix}
\usepackage{xcolor,colortbl}
\usepackage{hyperref}
\def\BibTeX{{\rm B\kern-.05em{\sc i\kern-.025em b}\kern-.08em
    T\kern-.1667em\lower.7ex\hbox{E}\kern-.125emX}}
\let\oldemptyset\emptyset

\begin{document}
\title{Search and Score-based Waterfall Auction Optimization}
%
%
\author{Dan Halbersberg \and
Matan Halevi \and
Moshe Salhov}
\authorrunning{Halbersberg et al.}
%
\institute{Playtika Ltd.\\
Hahoshlim St. 8, Herzliya, Israel\\
\email{danh@playtika.com | matanh@playtika.com | Moshesa@playtika.com}}
\maketitle              
\begin{abstract}
	Online advertising is a major source of income for many online companies. One common approach is to sell online advertisements via waterfall auctions, through which a publisher makes sequential price offers to ad networks.~The publisher controls the order and prices of the waterfall in an attempt to maximize his revenue. In this work, we propose a methodology to learn a waterfall strategy from historical data by wisely searching in the space of possible waterfalls and selecting the one leading to the highest revenues.~The contribution of this work is twofold; First, we propose a novel method to estimate the valuation distribution of each user, with respect to each ad network. Second, we utilize the valuation matrix to score our candidate waterfalls as part of a procedure that iteratively searches in local neighborhoods.~Our framework guarantees that the waterfall revenue improves between iterations ultimately converging into a local optimum. Real-world demonstrations are provided to show that the proposed method improves the total revenue of real-world waterfalls, as compared to manual expert optimization. Finally, the code and the data are available \href{https://github.com/PlaytikaResearch/public_waterfall2}{\textcolor{blue}{here}}.

\end{abstract}
\section{Introduction}

Online advertisements serve as a major source of income, for many online companies~\cite{muthukrishnan2009ad}. Whenever a user surfs on a website or utilizes a mobile app, an advertisement real-estate, known as \textit{ad-slots} are allocated. Each ad-slot is populated by a relevant advertisement. The slot owner is called a \textit{publisher} and the advertisement owner is called a \textit{supplier}. The publisher sells ad-slots to suppliers via \textit{ad networks} such as Facebook, Google, etc. Interactions between publishers and suppliers take place through real-time bidding auctions. 

The publisher's goal is to maximize the selling price of each ad-slot. There are several approaches to choosing a particular ad network for a given slot. One common approach is known as the waterfall strategy~\cite{afshar2019decision,busch2016programmatic,chakraborty2010approximation,despotakis2019first,kveton2019waterfall,wang2016display}. A waterfall is a list of instances, where each \textit{instance} belongs to a specific ad network and is associated to a specific price~\cite{kveton2019waterfall}. For each ad-slot, the ad networks are sequentially approached, according to a pre-configured list of instances. Each ad network can accept or reject to buy the slot for the given pre-defined instance price. If the ad network rejects the price, the slot is offered to the next instance in line until an ad network accepts the terms. Since the waterfall is predefined by the publisher, optimally determine the strategy, i.e., the order and pricing configuration of all instances in the waterfall, remains a significant challenge~\cite{afshar2019reinforcement,afshar2021reward}.

To empower the user experience, this bidding process must be completed in real-time in order to empower user experience ~\cite{ting2018maximizing,zhang2014optimal}. Therefore, it is important to find the best strategy, such that the timing of the last approached instance will not breach the time constraint. Additionally, the supplier limits the number of instances, to minimize the auction overhead~\cite{afshar2019reinforcement}.

Many publishers decide on the ordering of ad networks based on human experience, trial and error, or other similar inefficient methods~\cite{afshar2019reinforcement}. Besides for the fact that these methods cannot guarantee an optimal strategy, it is challenging to scale them for large publishers who need to manage many waterfalls for different platforms, operation systems, countries, etc. In recent years, growing attention has focused on automating and optimizing this process, in order to increase the revenue. An online learning algorithm was proposed by~\cite{kveton2019waterfall} to solve this problem, and other reinforcement methods were proposed~\cite{afshar2021reward,afshar2019decision,afshar2019reinforcement,wang2016display} as well. These methods focus on predicting the ad network's pricing strategy. In many real-world cases, the waterfall strategy is repeatedly operating over the same users and over several online sales events. This data stream can be accumulated and utilized to further improve any waterfall strategy.

In this study, we propose a novel approach that utilizes user-accumulated data that is measured during multiple visits to the publisher's web site or app. More explicitly, we suggest to use this information to estimate the perceived valuation of each user, by each ad network. These personalized valuations inform our hypothesis regarding auction events via simulations. The ability to simulate the actual effect of the bidding process allows us to design an efficient local search strategy for designing a locally optimal waterfall strategy. Applying the optimal waterfall strategy allows the publisher to not only gain more profits from the auction, but also to automate the process and adapt to temporal changes in user valuation, which reduces the overhead from marketing experts.

The rest of the paper is organized as follows. Section~\ref{background} provides the relevant background and surveys related works. In Section~\ref{Method}, we present the proposed framework and detail its characterizations. Furthermore, Section~\ref{Results} demonstrates the application of the suggested method using both simulated and real-world data. Finally, in Section~\ref{Summary} we provide our conclusions.

\section{Background and related works}
\label{background}

\begin{figure}[t!]
	\centering
	\includegraphics[width=0.5\textwidth]{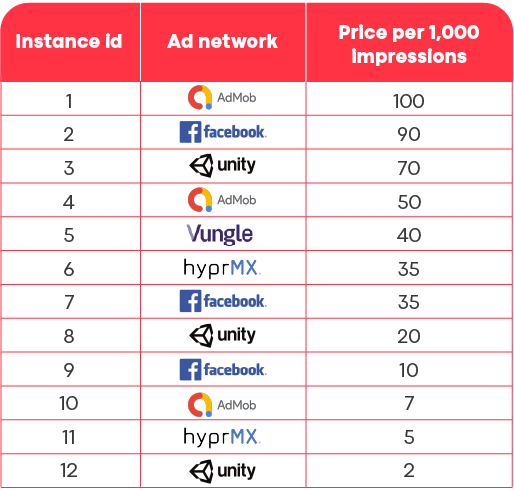}
	\caption{An example of a waterfall with twelve instances of five ad networks.}
	\label{WaterfallExample}
\end{figure}

A waterfall, $W$, is an ordered list of $r$ instances. The $i^{th}$ instance $W_i$, where $1 \le i \le r$ is associated to a specific ad network for a given price, $p_i \in \{0,...,M\}$, where $M$ is the maximal price allowed by the ad network~\cite{kveton2019waterfall}. Thus, each ad network in the waterfall may have several instances. For example, in Figure~\ref{WaterfallExample}, the ad network $AdMob$ has three instances $1, 4$, and $10$ for $\$100, \$50$, and $\$7$, respectively. The waterfall strategy is not tailored to a specific user valuation, but rather is designed to be optimal for the entire user population (i.e., all users will run through the same waterfall).

Most of the works in this field are based on concepts pertaining to reinforcement learning~\cite{afshar2019reinforcement}. One such algorithm~\cite{kveton2019waterfall} is a multi-armed bandits algorithm. The idea of the algorithm is to learn the valuation distribution of each ad network. For that, the publisher adaptively chooses waterfall strategy, receives feedback (accept/reject), and evaluates the performance using a regret function. However, this online algorithm has several restrictive assumptions, such as: each ad network has a single instance and that the valuations are unique per ad network, yet equal for all users. Similar assumptions were also made by other researchers~\cite{afshar2019reinforcement}.
Recently,~\cite{afshar2019decision} proposed to utilize the ad-request information to learn a model that predicts the probability of an ad network to buy the ad-slot for the given price. The outputs are then fed into a Monte-Carlo algorithm, which optimizes the state-action values, accordingly. However, proposing a policy for each ad-request (even in the case that they are grouped by their commonality) is practically impossible.

The waterfall optimization can be defined as a local search problem, where the task is to maximize the total waterfall strategy revenue over all users. Search and score (S\&S) is a heuristic method that belongs to a family of local search algorithms~\cite{hoos2004stochastic}. Methods following this heuristic aim to solve computationally hard optimization problems~\cite{battiti2008reactive,hoos2004stochastic}. One such method is known as hill climbing~\cite{russell2002artificial}; an iterative algorithm is initialized by an arbitrary solution and then attempts to find a better solution candidate by making a local change to the best solution so far. The best solution candidate is adopted by the algorithm, which is then followed by a similar evaluation of any incremental change to that selected solution, in the next iteration. The algorithm terminates once the improvement is negligible.

A major limitation of the hill climbing procedure is that it can converge into a local minimum.~This limitation is heavily dependent on the search starting point. To overcome this limitation, another heuristic search known as the Monte Carlo Tree Search (MCTS) was proposed~\cite{coulom2006efficient}. As opposed to the hill climbing procedure, the MCTS searches for the most promising next solution candidate in a decision-making problem, combining the precision of tree search with the generality of random simulation. Algorithms following the MCTS approach adopt, in each iteration, the change with the greatest potential with respect to future iterations. 
One relevant implementation of a hill climbing procedure, which inspired our proposed S\&S--based waterfall optimization algorithm, is the well-known K2 algorithm~\cite{cooper1992bayesian}. The K2 is a heuristic search algorithm for learning the structure of a Bayesian network that best fits the data. The algorithm starts with a random graph, and considers all local neighbor graphs, at each iteration. A neighbor graph is defined as an equal graph with a single change that can be: edge addition, deletion or reversal. By likening the waterfall $W$ to a serial graph, we can equate a node in the graph to an instance in the waterfall and edges to the waterfall order. Using the estimated valuation matrix $B$, a procedure, similar to that of the K2, can be designed over the space of all valid waterfall graphs. In the following section, we will describe this local search procedure in detail.

\section{Proposed method}
\label{Method}

To utilize the fact that users are frequently visiting the publisher's website or app, we propose a two-stage framework; First, estimate valuation matrix, $B \in \mathcal{R}^{U \times K}$ from historical data, where $U$ is the number of distinct users and $K$ is the number of ad networks. $B$ holds the perceived value of each user by each ad network (Section~\ref{Estimate}). Second, search for the optimal waterfall by simulating auction events utilizing $B$, which can approximate the revenue effect of order and pricing changes in the waterfall. More explicitly, in the second stage, our proposed framework uses the valuation matrix in an iterative manner (Section~\ref{Search}). In this way, we are able to define the problem as a local search problem over the space of valid waterfalls.

The main contributions of the proposed method are: 1) modeling user pricing per ad network, by learning each user's Beta distribution parameters, given their respective historical pricing data; 2) minimizing the information requirement by explicitly utilizing sales events, while implicitly utilizing rejected bid information from the sales data. The two contributions rely on the fact that the advertisement process is ongoing and most user ad-slots were sold several times in the past.

\subsection{Estimate the Valuation Matrix}
\label{Estimate}

The valuation matrix ($B$) is a key component in our S\&S--based waterfall optimization algorithm as the entire search procedure depends on $B$ to simulate auction events.~If $B$ is wrong, the results would be misleading.~One simple approach to estimating $B$ is to take the average sell price of each user, per ad network. However, the main limitations of this metric are: 1) it will generate a deterministic value that is less suitable for simulation purposes; and 2) in many cases, the average is not a good representative of the user valuation. To overcome these limitations, one can suggest to replace the simple average with a normal distribution estimation, and then during the search phase, sample from this distribution. Nevertheless, normal distribution does not necessarily fit the user valuation distribution, and therefore, we propose to use the Beta distribution, which allows for a more flexible representation of the user's valuation. This distribution was also found appropriate by other researchers~\cite{chou2017improving, kveton2019waterfall}. Algorithm~\ref{AlgBeta} loops over each user ($u, \quad 1 \le u \le U$) and collects the user data (vector $V_{u,k}$) per ad network ($k, \quad 1 \le k \le K$) for the beta estimation ($B_{u,k}$). This is our training data. However, if no data is available, i.e., the user was never sold to that particular ad network, the algorithm tries two estimation methods; 1) if the user was sold in the past to at least $a$ (in our experiments we use $a=3$) other ad networks, it uses their data for the estimation; or 2) if (sufficient) other ad networks data are not available, it uses the global beta distribution parameters of the specific ad network in question. The output of the Algorithm is the valuation matrix $B_{u,k}$ of dimension $U \times K$, where $U$ is the number of distinct users and $K$ is the number of ad networks. In our case, the main advantage of the Beta distribution over, for example, the simple mean value, is that it generalizes well to a stochastic process, allowing us to represent different types of users. For example, some users could have a Poisson-like distribution while others can be normal, exponential, etc. In general, the beta distribution is appropriate when the true probability distribution is unknown~\cite{johnson1995continuous}.
\begin{algorithm}[t]
	\caption{Estimate the valuation matrix}
	\label{AlgBeta}
	\begin{algorithmic}[1]
		\State \textbf{Input:} Dataset $D$
		\State \textbf{Output:} Valuation matrix $B$  \\
		\For{each User $u$ and ad network $k$}
		\State $V_{u,k}=$ collect all past sell events of $u$ to $k$ from $D$
		\If {$V_{u,k} \neq \oldemptyset$ }
		\State $B_{u,k}=beta.fit(V_{u,k})$
		\ElsIf {$u$ was sold to at least $a$ ad networks $z \neq k$}
		\Comment{Imputation needed}
		\State $V_{u,k}=$ collect all past sell events of $u$ to $\forall z \neq k$
		\State $B_{u,k}=beta.fit(V_{u,k})$
		\Else
		\State $B_{u,k}=B_k$ \Comment{Use global parameters of $k$}
		\EndIf
		\EndFor
	\end{algorithmic}
\end{algorithm}

Another method that can be used to estimate the valuation matrix is via a classifier (e.g., CATboost, Neural Network, etc.).~Following~\cite{afshar2019reinforcement}, we propose to utilize historical data to predict the valuation of each user, and more explicitly, to train classifiers per ad network. To achieve this goal, one can use information from past auction events to correlate between the dependent variable (auction price) and other available independent variables such as: time, geography, demographic, device information, in-app activities, etc. Once such classifiers are trained, we may use them to fill-in valuation matrix, $B_{u,k}$.

\textbf{Evaluate the valuation matrix estimation in terms of accuracy}: It is crucial to validate the capability of the estimated Beta distributions in $B$ to accurately generate pricing predictions, before moving forward to search for the best fitted waterfall, using that matrix. The $B$-based pricing predictions ability is a key component in our framework. Large prediction errors will result in a misleading waterfall strategy that will produce reduced revenues, when run over the actual user population.

\begin{algorithm}[t]
	\caption{Run users in a waterfall}
	\label{AlgRunUsers}
	\begin{algorithmic}[1]
		\State \textbf{Input:} Valuation matrix $B$, Waterfall $W$
		\State \textbf{Output:} $q'$ \\
		\State \textbf{Initialization:} 
		\For{each instance $W_i$ in the waterfall $W$}
		\State $q'_i=0$
		\EndFor
		\State \textbf{Start:} 
		\For{each User $u$}
		\State $i=0$ \Comment{Go over the waterfall from top to bottom}
		\While{$i \leq r$}
		\State $k$ = ad network of $W_i$
		\State $P_{u,k}=$ Sample a value from $B_{u,k}$
		\If {$P_{u,k} > p_i$ }
		\State $q'_i+=1$
		\State Break
		\EndIf
		\State $i+=1$
		\EndWhile
		\EndFor
	\end{algorithmic}
\end{algorithm}
\begin{algorithm}[t]
	\caption{Evaluate the valuation matrix}
	\label{AlgEval}
	\begin{algorithmic}[1]
		\State \textbf{Input:} Valuation matrix $B$, Waterfall $W$
		\State \textbf{Output:} Similarity score \\
		\State $q'=$ Run users in the waterfall based on their valuation matrix \Comment{Call Algorithm~\ref{AlgRunUsers}}
		\State $Score = 0$
		\For{each instance $W_i$ in the waterfall $W$}
		\State $Score += \frac{|q'_i-q_i|}{q_i} \times We_i$
		\EndFor
	\end{algorithmic}
\end{algorithm}

To evaluate $B$-based prediction accuracy, we propose to generate sale predictions for validation data that is accumulated similarly to the training data. Each sale event is an advertisement (sometimes called an \textit{impression}) and has corresponding sale pricing. Acceptable $B$ will predict the number of impressions with good accuracy, as compared to the given validation impressions data. The performance evaluation process is detailed in  Algorithms~\ref{AlgRunUsers} and~\ref{AlgEval}. The goal of Algorithm~\ref{AlgRunUsers} is to predict how each ad network instance in a predefined waterfall strategy would behave, given $B$. Let $q \in \mathcal{Z}^{+r}$ be the measured vector of the actual outcomes of a waterfall strategy in the form of the number of impressions per ad network instance and $q_i$ an impression assignment of a specific instance. The output of Algorithm~\ref{AlgRunUsers} is a vector, $q'$, which holds the number of predicted impressions of each instance in the waterfall. Algorithm~\ref{AlgEval} takes $W$ and its corresponding impressions, $q$, as input. Additionally, Algorithm~\ref{AlgEval} compares $q$ and $q'$ computed in Algorithm~\ref{AlgRunUsers}, based on Equation~\ref{EqCompScore}.

\begin{equation} \label{EqCompScore}
	Score = \sum_{i=1}^{r}\frac{|q'_i-q_i|}{q_i} \times We_i,
\end{equation}
where $q'_i$ and $q_i$ are the simulated and real number of impressions in the ${i}^{th}$ instance, respectively, $r$ is the length of the waterfall, and $We_i$ is the weighted revenue of the ${i}^{th}$ instance: $We_i=\frac{Revenue(W_i)}{\sum{Revenue(W_i)}}$. If this score is low enough, we say that $B$ represents the true value of the users, as perceived by the different ad networks.

\subsection{The search procedure} \label{Search}

First, we define the score function that is used to compare waterfalls, as part of the S\&S procedure:
\begin{equation} \label{EqRevenue}
	Revenue(W)=\sum_{i=1}^{r}q_i \times p_i.
\end{equation}

\begin{figure}[b!]
	\centering
	\includegraphics[width=0.5\textwidth]{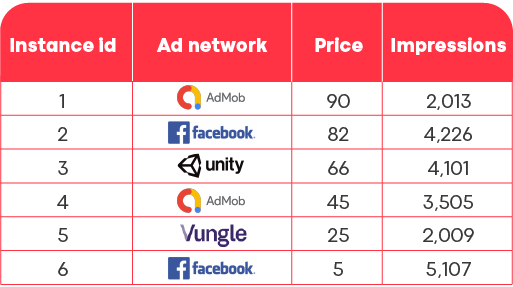}
	\caption{An example of a waterfall with six instances and their associate daily number of impressions for a given date. The total daily revenue of the waterfall is \$1,032 since the prices in the waterfall are for batches of 1,000 users.}
	\label{WaterfallImpressionsExample}
\end{figure}

The revenue of each candidate waterfall is calculated using Algorithm~\ref{AlgRunUsers}. That is, once the algorithm runs all the users through the candidate waterfall, the number of impressions of each instance, $q'_i$, are updated. Figure~\ref{WaterfallImpressionsExample} shows an example output of Algorithm~\ref{AlgRunUsers}, where the last column is $q'$. Using Equation~\ref{EqRevenue}, we can sum up the multiplication of the price and the number of impressions to get the total revenue of the candidate waterfall.

\begin{figure}[t!]
	\centering
	\includegraphics[width=0.75\textwidth]{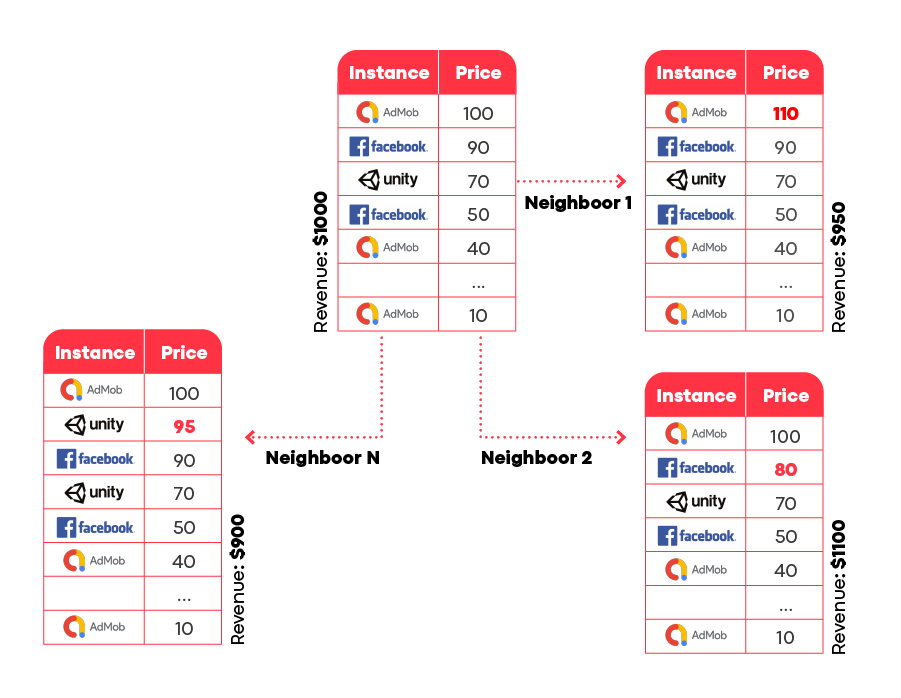}
	\caption{An illustration of a single iteration in the S\&S-based waterfall optimization algorithm. In each neighbor waterfall the changed price is marked in red.}
	\label{FigSandS}
\end{figure}

Next, our search procedure considers all local neighbors (i.e., candidate waterfalls) in each iteration, where a local neighbor is defined as the current waterfall, except for a single change that can be: 1) instance addition, 2) instance removal, or 3) a changed instance price (increase or decrease). To narrow-down the search space, we restrict the prices to be discrete (however, one can choose other quantiles such as 50 cents, 10 cents, etc.). By scoring each neighbor waterfall, the algorithm can evaluate how the waterfall would be affected by each specific incremental change. In each iteration, the algorithm adopts the change leading to the highest improvement in revenue. Figure~\ref{FigSandS} illustrates an example of a waterfall with a revenue of $\$1,000$ and three of its neighbor waterfalls. The second neighbor has the highest revenue, and therefore, is selected as the incremental change to the next iteration. 

\begin{algorithm}[t]
	\caption{Search and score procedure}
	\label{AlgSandS}
	\begin{algorithmic}[1]
		\State \textbf{Input:} Initial waterfall $W^0$
		\Comment{This could be the current existing waterfall}
		\State \qquad \quad Valuation matrix $B$
		\State \qquad \quad Max\_iter
		\State \qquad \quad $\epsilon$
		\State \textbf{Output:} Waterfall $W$  \\
		\State \textbf{Initialization:} 
		\State $q'=$ Run users in the waterfall $W^0$ based on their valuation matrix $B$ 		\Comment{Call Algorithm~\ref{AlgRunUsers}}
		\State Optimal\_revenue = Revenue($W^0$)
		\State Convergence = True
		\State $W$ = $W^0$
		\State Iter $=0$
		\State \textbf{Start:} 
		\While{Convergence}
		\State Iter += 1
		\State Prev\_revenue = Optimal\_revenue
		\State neighbors = list of neighbors of $W$
		\For{each neighbor $g$ in neighbors}
		\State {Run the users ($B$) in $g$} \Comment{Call Algorithm~\ref{AlgRunUsers}}
		\State Curr\_revenue = Revenue($g$)
		\If {Curr\_revenue $>$ Prev\_revenue}
		\State Prev\_revenue = Curr\_revenue
		\State $W$ = $g$
		\EndIf
		\EndFor
		\If {Iter==Max\_iter \textbf{or} Prev\_revenue - Optimal\_revenue $<\epsilon$}
		\State Convergence = False
		\State Optimal\_revenue = Prev\_revenue
		\EndIf
		\EndWhile
	\end{algorithmic}
\end{algorithm}

Following, we present our proposed S\&S--based waterfall auction optimization algorithm (Algorithm~\ref{AlgSandS}). The algorithm takes (lines 1--5) an initial waterfall ($W^0$), a valuation matrix ($B$), the maximal number of iterations ($Max\_iter$) and a threshold ($\epsilon$) for the stopping condition as inputs. The initial waterfall can be an empty/random waterfall, or an existing/human expert waterfall. After initialization (lines 9--13) is complete, the algorithm iterates, taking into consideration all local changes (line 17) and adopting the one with highest revenue (line 22), for each iteration. Finally, it terminates once the maximal number of iterations is reached or the difference in revenue between two successive iterations is determined to be lower than $\epsilon$ (line 26).

Our waterfall auction optimization algorithm is based on the heuristic hill climbing method~\cite{hoos2004stochastic} that can neither guarantee reaching the global optimum nor converging to a local or the global optimum in a reasonable time frame. Since our proposed algorithms adopt a neighbor waterfall only if its revenue is strictly above the aforesaid threshold, then no cycles can exist within the search procedure, and thus, the algorithms must converge. The convergence rate, however, is problem-dependent~\cite{cooper1992bayesian}. Luckily, our target function is the total revenue of the waterfall strategy and the majority of the revenue comes from the higher section of the waterfall, where the prices are relatively high. Since the algorithm selects the change leading to the highest improvement in revenue for each iteration, it will first optimize the higher part of the waterfall, enabling the best solution to be reached relatively quickly. Therefore, one can restrict the number of iterations, as we show in our empirical evaluation in Section~\ref{Results}.

\textbf{The Monte Carlo tree search (MCTS):} Another search procedure we chose to apply as part of our waterfall auction optimization algorithm is a MCTS-like procedure. The goal of this algorithm is to expand the search space in order to avoid a local maximum, due to the non-convex nature of our score function. Algorithm~\ref{AlgMCTS} (see Appendix~\hyperref[App.A]{A}) trades between accuracy and complexity, as it evaluates more candidate waterfalls at the expense of a higher run-time. At each iteration, the algorithm will adopt the neighbor waterfall with the greatest revenue potential; not necessarily the one with the current highest revenue. All other components of the MCTS--based waterfall optimization algorithm are the same as the regular S\&S (i.e., Algorithms~\ref{AlgBeta}--\ref{AlgEval}).

\textbf{Complexity:} The computational complexity of our proposed methods is composed of two elements, corresponding to the two folds of our framework. The first, which dictates the complexity, is the estimations of $B$ per user and ad network. In total, the algorithm has to estimate $U \times K$ Beta distributions parameters, each from $x$ samples per user (past sell events), which depends on the data time-period ($x$ is monotonic with the data duration). The second, less dominant element of our framework is the hill climbing search. In this stage, the algorithm is bounded by $O(U \times r)$ for each candidate waterfall, since in the worst case, each user runs through the entire waterfall. The number of candidate waterfalls at each of the $it$ iterations is bounded by $O(3r+MK)$, and thus, the second element of our framework is bounded by $O[it \times n \times r \times (3r+MK)]$.

The number of possible waterfalls is $(M \times K)^r$, where $M$ is the number of unique prices ($M$ is also the maximal price in our discrete case). Although the number of possible iterations is bounded by the number of possible waterfalls, in practice, $it$ is restricted to a value between $10-50$. This is because the optimization is mainly affected by changes in the higher section of the waterfall (as described in Section~\ref{Search}), and thus, we achieve most of the improvement in revenue at an early stage of the search process.

Moreover, both parts of the computations (i.e., estimating $B$ and the hill climbing search) could easily be parallelized: 1) estimating $B$ could be parallelized by distributing the data by users, and; 2) as part of the search procedure, one could evaluate at each iteration the candidate waterfalls in parallel and by that reduce the algorithm run-time to $O(it \times U \times r)$. Also, note that $it$ is small and that $O(U \times r)$ refers to the worst case scenario.~In practice, assuming, for example, that the impressions are distributed uniformly across the waterfall, this bound is reduced by half, as Equation~\ref{EqBound} shows regarding the number of requests to ad networks:

\begin{equation} \label{EqBound}
	\#\_of\_requests=\sum_{i=1}^{r}i \times \frac{U}{r} = \frac{r (r+1)}{2} \times \frac{U}{r} = \frac{U(r+1)}{2}.
\end{equation}

Finally, in terms of the data we need to store, our proposed algorithms require only the successful events. This is a huge advantage over other reinforcement learning algorithms (e.g.,~\cite{afshar2019decision,kveton2019waterfall}), which require both accepted and rejected auction events. This allows us to reduce the volume of stored data by over $90\%$.

\section{Empirical evaluation}
\label{Results}
 
In this section, we report on our experiments with synthetic and real-world waterfalls. With respect to the latter, we experimented with four different waterfalls linked to different countries, to increase the variability of our results. To maintain confidentially, we will refer to them as $Waterfall_A-Waterfall_D$.~We compared the two variations of the search procedure; S\&S-- and MCTS--based waterfall optimization algorithms, to a human expert optimization based on total revenue and computational complexity. The waterfalls and their associated data can all be found online in our \href{https://github.com/PlaytikaResearch/public_waterfall2}{\textcolor{blue}{supplementary materials}}. The human expert optimization is actually the most common approach in the industry, and therefore, it can be used as a solid baseline.
 
\subsection{Synthetic data} \label{Synth}

\begin{figure}[b!]
    \centering
    \includegraphics[width=0.39\textwidth]{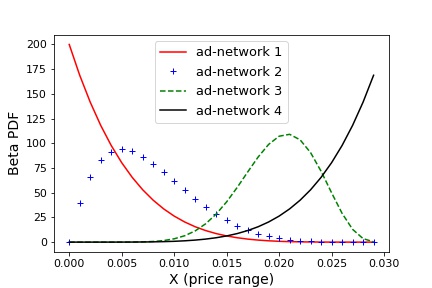}
    \caption{The beta distributions used to sample the synthetic valuations.}
    \label{FigSyntBeta}
\end{figure}
In this section, we report on our experiment with synthetic data.~The motivation for this experiment was twofold: 1) to show that our proposed S\&S--based algorithm may converge to the optimal solution regardless of the initialization, using a toy example, and; 2) to show that the simple S\&S--based algorithm does not fall behind the MCTS--based algorithm, in terms of accuracy, while it convergences significantly faster.~To demonstrate these advantages, we selected four ad networks and synthetically sampled $4 \times 10^5$ users, where the beta distribution parameters per ad network were: $Beta(\alpha=1,\beta=6), Beta(2,6), Beta(10,5)$, and $Beta(6,1)$. We selected these parameters so that the distributions would substantially overlap each other (as can be seen in Figure~\ref{FigSyntBeta}). In addition, we initialized the algorithms with five different waterfalls, i.e., different in the order and the prices of the instances, to show that the convergence was not random.

\begin{figure*}[t]
	\centering
	\begin{subfigure}[t]{0.47\textwidth}
		\centering
		\includegraphics[width=\textwidth]{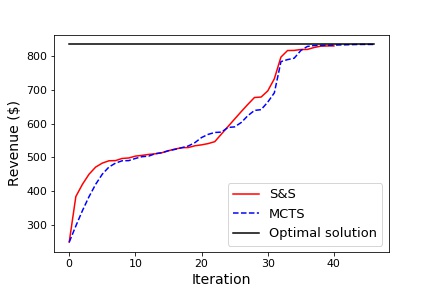}
		\caption{}
		\label{FigSyntRev}
	\end{subfigure}
	\hfill
	\begin{subfigure}[t]{0.47\textwidth}
		\centering
		\includegraphics[width=\textwidth]{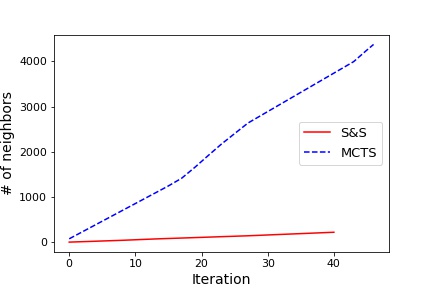}
		\caption{}
		\label{FigSynNei}
	\end{subfigure}
	\caption{a) The learning curve measured in revenue (\$) for the S\&S-- (solid red line) and MCTS--based (dashed blue line) algorithms comparing to the optimal solution that was found using exhaustive search and b) The cumulative number of neighbors (candidate waterfalls) examined by the S\&S-- (solid red line) and MCTS--based (dashed blue line) algorithms, with respect to the learning curve in Figure~\ref{FigSyntRev}.}
\end{figure*}

Figure~\ref{FigSyntRev} shows the learning curve of the two search algorithms for an empty waterfall initialization. The optimal solution (\$835.4) was calculated using an exhaustive search over all possible solutions (total of 30 discrete prices and 4 ad networks, generating $4! \times 30^4$ candidate waterfalls). It can be seen that both S\&S-- and MCTS--based waterfall optimization algorithms converge to a solution that is close to the optimal one (\$830.9 and \$835.4 for the S\&S and MCTS, respectively). Although the S\&S--based algorithm did not achieve as optimal a solution as the MCTS--based algorithm, their revenue is close enough ($\frac{830.9}{835.4}=0.995$). In addition, the fact that they both have a similar number of iterations ($\approx 40$) is deceiving, as the number of examined neighbors is significantly lower for the S\&S--based algorithm (as can be seen in Figure~\ref{FigSynNei}). The S\&S--based algorithm simulated only 200 waterfalls, as opposed to the over 4,000 waterfalls simulated by the MCTS--based algorithm. Therefore, concerning the trade off accuracy-runtime, this experiment demonstrates the S\&S--based algorithm might be superior to the MCTS--based algorithm, with respect to both accuracy and complexity.

\setlength{\textfloatsep}{15pt}

Table~\ref{TblSynRes} shows the results of the S\&S-- and MCTS--based waterfall optimization algorithms for five initializations: 1) the true order, but with different prices, 2) empty waterfalls, 3) all prices being equal to the average valuation, and 4-5) in the opposite order and with different prices. It can be seen that, except for the last two initializations, the MCTS--based algorithm learned the optimal waterfall, while the S\&S--based algorithm did not recover the optimal waterfall for any of the initializations. Nevertheless, the S\&S--based algorithm is only $0.6\%$ lower than the MCTS--based algorithm on average. That said, it is $\approx 20$ quicker, as measured by the number of neighbors evaluated during the learning. Therefore, we can conclude that for small non-complex waterfall optimization problems, the S\&S--based algorithm can almost achieve the global optimum and it is monotonic with the MCTS--based algorithm, but faster in several orders of magnitudes.

\begin{table}[t!]
	\begin{center}
		\begin{tabular}{cccccc}
			        & Init\_1 & Init\_2 & Init\_3 & Init\_4 & Init\_5				 \\\hline
			S\&S    & 833.9 &  830.9    & 822.5 & 829.5 & 827.9 	 \\
			MCTS    & 835.4 &	835.4   & 835.4 & 831.4 &	832.4	 \\
			Difference (\%) & 0.2 &	0.5 & 1.6   & 0.2   &	0.5	
		\end{tabular}
	\end{center}
	\caption{The revenue(\$) achieved by the two search algorithms for five initializations, where the optimal solution is \$835.4}
	\label{TblSynRes}
\end{table}

\subsection{The real-world auction data} \label{Data}

The data used in our analyses were collected from four different real waterfalls. The "raw-data" used in our experiments are hourly aggregated, per user and ad network. The given data were randomly sampled from a cohort of $60$ days which include $355,905$ users and $27,615,614$ advertisements (an average of $78$ advertisements per user). The data processing is described in detail, in Table~\ref{tab:dataPrep} in Appendix~\hyperref[App.B]{B}.

Next, we describe the experiment methodology. The given data includes $60$ days of waterfall sales and strategy measurements. For each validation day number $d$, we use its past $30$ days as training data. The result is training set $X_t$ and a corresponding validation data $X_v$. Given the data train-validation split we used Algorithm~\ref{AlgBeta} on $X_t$ to learn the beta distributions in $B$. Given $B$, we predicted the outcome of applying the waterfall strategy $W$ on samples from $B$ using Algorithm~\ref{AlgRunUsers}. Furthermore, we generated the corresponding impressions vector $q'$. This vector was compared to the given impressions vector $q$ of the waterfall from day $d$ using Equation~\ref{EqCompScore}. Notice, that $q'$ was estimated based on data ended in day $d-1$, while $q$ refers to day $d$. This is to prevent over-fit and to demonstrate the prediction capability of $B$.

Since a sale event can take place only if the user valuation is larger than the instance price, we need to increase the values sampled from $B$ by a small constant $\epsilon$. To enable this to occur, we define a coefficient vector $\zeta$ that includes one coefficient per ad network. We learn the $\zeta$ that minimize Equation~\ref{EqCompScore}. However, optimizing all coefficients at once is practically unfeasible, since a change in one coefficient requires to re-predict the  waterfall outcome and re-evaluate $B$, based on Equation~\ref{EqCompScore}. Therefore, we suggest to learn the coefficients in a coordinate descent methodology~\cite{bertsekas2015convex}. That is, randomly selecting an ad network and optimizing its coefficient using a grid search. Then, move to the next ad network and continue the optimization in a round-robin manner, until the improvement in Equation~\ref{EqCompScore} is negligible. This procedure allows to loop over the ad networks several times. Finally, after $B$ is estimated, run algorithm~\ref{AlgSandS} and~\ref{AlgMCTS} to find the converged waterfall. We use the current real waterfalls as the initial waterfall inputs for the learning algorithms. This enable us to exploit the human expert knowledge.

\begin{figure*}[t]
	\centering
	\begin{subfigure}[b]{0.42\textwidth}
		\centering
		\includegraphics[width=\textwidth]{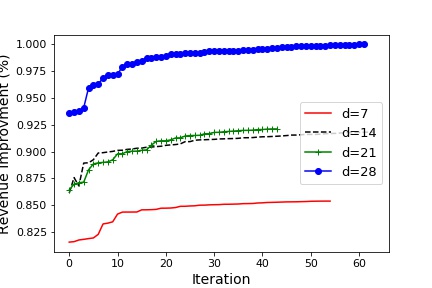}
		\caption{$Waterfall_A$}
		\label{FigRwWaterfall1}
	\end{subfigure}
	\hfill
	\begin{subfigure}[b]{0.42\textwidth}
		\centering
		\includegraphics[width=\textwidth]{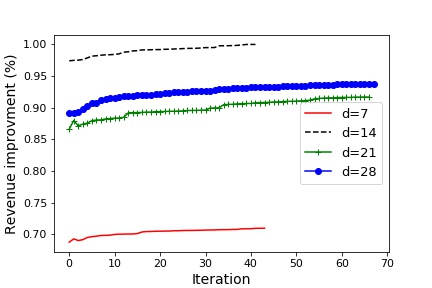}
		\caption{$Waterfall_B$}
		\label{FigRwWaterfall2}
	\end{subfigure}
	\\
	\begin{subfigure}[b]{0.42\textwidth}
		\centering
		\includegraphics[width=\textwidth]{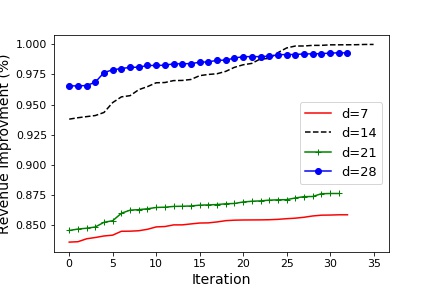}
		\caption{$Waterfall_C$}
		\label{FigRwWaterfall3}
	\end{subfigure}
	\hfill
	\begin{subfigure}[b]{0.42\textwidth}
		\centering
		\includegraphics[width=\textwidth]{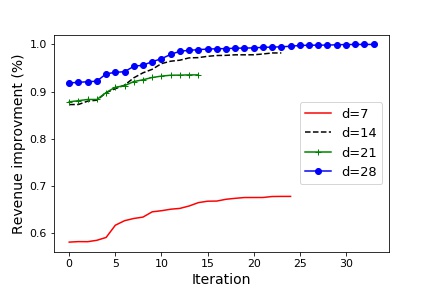}
		\caption{$Waterfall_D$}
		\label{FigRwWaterfall4}
	\end{subfigure}
	\caption{The learning curve, measured as the improvement in revenue, for the four real-world waterfalls ($A-D$). Each color represents a point in time ($d$) to which the algorithm was applied.}
	\label{FigMainResults}
\end{figure*}

Table~\ref{TblFitness} shows the valuation matrix fitness to each waterfall, as measured by Equation~\ref{EqCompScore}. The values in the table are all positive, but are not bounded i.e., $[0-\infty]$. There is a trade-off between an accurate valuation matrix and a generalized one. One could define the valuation matrix as the true absolute sales prices of each user that will generate $q$, but this will result in over-fit that will mislead the waterfall strategy over new data. Thus, the proposed valuation matrix, is a probabilistic estimation of longer time period, which will generalize well, and allows the algorithm to explore and exploit the data. Table~\ref{TblFitness} demonstrates that, for $Waterfall_A$, the valuation matrix is the most accurate with a mean error of $0.29$ and a corresponding std of $0.06$. This is the result of a larger number of ad sales per user in $Waterfall_A$, as compared to the other waterfalls.
\begin{table}[t!]
    \begin{center}
    	\begin{tabular}{lccccc}
    		& $d=7$ & $d=14$ & $d=21$ & $d=28$ & mean (std)	    \\\hline
    		$Waterfall_A$ & 0.30 & 0.37 & 0.27 & 0.23 & 0.29 (0.06) \\
    		$Waterfall_B$ & 0.62 & 0.78 & 0.70 & 0.66 & 0.69 (0.07) \\
    		$Waterfall_C$ & 0.46 & 0.54 & 0.48 & 0.54 & 0.51 (0.04) \\
    		$Waterfall_D$ & 0.74 & 0.61 & 0.80 & 0.74 & 0.72 (0.08)
    	\end{tabular}
    \end{center}
    \caption{The valuation matrix fitness for four waterfalls and four time-points ($d$) measured by mean absolute error (low is better).}
    \label{TblFitness}
\end{table}

We will now present the results of our proposed search-based waterfall optimization algorithms. Figure~\ref{FigMainResults} shows the learning curves of the S\&S--based waterfall optimization algorithm for the four waterfalls in four time points during the experiment. In general, the algorithm converges after 10-20 iterations. Also, it can be seen that $Waterfall_A$ (Figure~\ref{FigRwWaterfall1}) has the fastest learning curve, as compared to the other waterfalls. This is due to the origin of the users in that particular waterfall. Furthermore, it can bee seen that the results for $d=7$ (red solid line) are always the lowest. This is due to the environment's characteristics in that period of time. Conversely, for $d=28$ (blue dot line) the results are always the highest (except for $Waterfall_B$).

\setlength{\textfloatsep}{15pt}

Table~\ref{TblRes} summarizes the results of the two search--based waterfall optimization algorithms, as compared to the human--based performances for the four real-world waterfalls. The human expert had manually inspected the original waterfalls at each time point, according to their number of impressions, prices, and ad network capacity, and following, recommendations for the changed waterfalls were evaluated in terms of total revenue (recall that this is the most common approach in the industry). The table reveals that human experts had the lowest improvement in revenue, while the MCTS had the highest average improvement rate. Although the MCTS was superior to the S\&S-based waterfall optimization algorithm for three out of four real waterfalls, this came at the expense of a much higher run-time: $20-25$ times higher. The run-time results could be seen in the last two columns of Table~\ref{TblRes}. Note that we did not parallelize the algorithms since a run-time of $\approx 1$ hour is reasonable for an off-line algorithm that usually runs once a day (for a more complex system that could be scaled).

\begin{table}[t!]
	\begin{center}
    	\begin{tabular}{lccccc}
			& \begin{tabular}{@{}c@{}}Human expert \\ (\%)\end{tabular} & \begin{tabular}{@{}c@{}}S\&S-based \\ (\%)\end{tabular} & \begin{tabular}{@{}c@{}}MCTS-based \\ (\%)\end{tabular} & \begin{tabular}{@{}c@{}}S\&S-based \\ (hours)\end{tabular} & \begin{tabular}{@{}c@{}}MCTS-based \\ (hours)\end{tabular}	  \\\hline
			$Waterfall_A$ & 0.6 & 6 & 5 & 1.2 & 27.0  \\
			$Waterfall_B$ & 0.6 & 4 & 4 & 1.8 & 47.4 \\
			$Waterfall_C$ & 0.1 & 4 & 5 & 0.8 & 17.1 \\
			$Waterfall_D$ & 0.2 & 11 & 12 & 0.5 & 11.6
    	\end{tabular}
	\end{center}
	\caption{The revenue improvement (\%) and run-time complexity (measured in hours), as compared to the baseline (current revenue) for the four waterfalls, as well as to the two search-based optimization algorithms and human expert.}
	\label{TblRes}
\end{table}

Finally, we experimented our algorithms with empty and a random waterfall initializations. First, we revealed that the empty initialization yield the worst results. The main reason is that the algorithm terminates too fast, and thus, converges to a poor local maximum. Second, we found that the random initialization was also inferior to the human expert initialization, which make sense, since the human expert knowledge is valuable.

\section{Summary}
\label{Summary}

In this study, we suggest a framework to learn about waterfalls from historical data. The settings in our proposed framework are: offline learning, multiple instances per ad network, unknown number of instances, and discrete prices. To the best of our knowledge, this is the first attempt to tackle the problem from the user's perspective. The main advantages of the proposed method are: it utilizes expert knowledge as an initial waterfall for the S\&S procedure, and; the proposed method does not requires an online search step. In many cases, online measurements are not feasible. Finally, our method uses only the successful events, while implicitly utilizing rejected bid information. This allows to significantly reduce the volume of the stored data by $90\%$.

Future research should focus on 1) improving the valuation matrix estimation.~Two proposals in this direction could be to use a more general distribution, such as Dirichlet, or to incorporate the rejected events in the valuation estimation process; 2) to dispense with the discrete price assumption, and; 3) to investigate other algorithms for solving this optimization problem so as to find the globally optimal waterfall strategy.~This is crucial, since the proposed search-based waterfall optimization algorithms can only guarantee sub-optimal solutions.\newpage

\bibliographystyle{plain}
\bibliography{arxivbib}\newpage

\section{Appendix}
\label{Appendix}

{\textbf{Appendix A - The pseudo-code for the Monte Carlo tree search algorithm}}\label{App.A}\\

In this Appendix, we present the pseudo-code for the MCTS proposed algorithm. As opposed to the S\&S-based algorithm, here, the algorithm will adopt the neighbor waterfall with the greatest revenue potential at every iteration; not necessarily the one with the current highest revenue. From an algorithm perspective, the difference is that there are two \textit{for loops} in lines 18 and 20.

\begin{algorithm}[ht]
	\caption{Monte Carlo tree search procedure}
	\label{AlgMCTS}
	\begin{algorithmic}[1]	
		\State \textbf{Input:} Initial waterfall $W^0$
		\Comment{This could be the current existing waterfall}
		\State \qquad \quad Valuation matrix $B$
		\State \qquad \quad Max\_iter
		\State \qquad \quad $\epsilon$
		\State \textbf{Output:} Waterfall $W$  \\
		\State \textbf{Initialization:} 
		\State $q'=$ Run users in the waterfall $W^0$ based on their valuation matrix $B$ 		\Comment{Call Algorithm~\ref{AlgRunUsers}}
		\State Optimal\_revenue = Revenue($W^0$)
		\State Convergence = True
		\State $W$ = $W^0$
		\State Iter $=0$
		\State \textbf{Start:} 
		\While{Convergence}
		\State Iter += 1
		\State Prev\_revenue = Optimal\_revenue
		\State neighbors = list of neighbors of $W$
		\For{each neighbor $g$ in neighbors}
		\State grand\_neighbors = list of neighbors of $g$
		\For{each neighbor $gg$ in grand\_neighbors}
		\State {Run the users ($B$) in $gg$} \Comment{Call Algorithm~\ref{AlgRunUsers}}
		\State Curr\_revenue = Revenue($gg$)
		\If {Curr\_revenue $>$ Prev\_revenue}
		\State Prev\_revenue = Curr\_revenue
		\State $W$ = $g$
		\EndIf
		\EndFor
		\EndFor
		\If {Iter==Max\_iter \textbf{or} Prev\_revenue - Optimal\_revenue $<\epsilon$}
		\State Convergence = False
		\State Optimal\_revenue = Prev\_revenue
		\EndIf
		\EndWhile
	\end{algorithmic}
\end{algorithm}\newpage

\begin{flushleft}
{\textbf{Appendix B - An example for the data processing flow}}\label{App.B}
\end{flushleft}

Table~\ref{tab:dataPrep} describes the data processing from raw-data (Table~\ref{TblData}) into a valuation matrix (Table~\ref{TblBeta}), using Algorithm~\ref{AlgBeta}. Table~\ref{TblVect} is the output of row $5$ in Algorithm~\ref{AlgBeta}. For example, rows $1$ and $4$ that are marked in red-bold in Table~\ref{TblData} are the raw data of user \textit{'4421AB3'} and \textit{'G'} with a single impression each. These two rows are converted to a vector with (at least) the two entries \textit{'[0.02,0.19]'} that are marked with a red-bold box in Table~\ref{TblVect}, before the beta distribution parameters, \textit{$Beta(\alpha=0.93,\beta=10.99)$}, are estimated as marked in red-bold in Table~\ref{TblBeta}.

\begin{table}[ht]
	\begin{subtable}{1\textwidth}
		\centering
    		\begin{tabular}{c|c|c|c|c|c}
    			Date & Hour & ad network & User id & Impressions & Revenue		 \\\hline
    			\cellcolor{gray}01/01/2021 & \cellcolor{gray}19 & \cellcolor{gray}G 		& \cellcolor{gray}4421AB3 & \cellcolor{gray}1 & \cellcolor{gray}0.020	 			\\
    			01/01/2021 & 18 & F		    & 345ADB  & 2 & 0.022 				\\
    			01/01/2021 & 21 & U			& 12345AS & 1 & 0.015 			    \\
    			\cellcolor{gray}01/01/2021 & \cellcolor{gray}19 & \cellcolor{gray}G			& \cellcolor{gray}4421AB3 & \cellcolor{gray}1 & \cellcolor{gray}0.019  	 	 	   	\\
    			01/01/2021 & 18 & F	        & 345ADB  & 2 & 0.018 				\\
    			01/01/2021 & 22 & U		    & 4421AB3 & 1 & 0.015 				\\
    			01/01/2021 & 20 & G 		& 12345AS & 3 & 0.057  	 	 	   	\\
    			...				 &  & &  &  &   	 	 	   	    
    	    \end{tabular}
    	\caption{An example of the raw-data}
    	\label{TblData}
	\end{subtable}
	\bigskip
	\begin{subtable}{1\textwidth}
		\centering
    		\begin{tabular}{c|c|c|c}
    			User id     & G 				& F & ...		 							\\\hline
    			4421AB3  &  \cellcolor{gray}[0.020,0.019,...]	& [...] &	 										  	 \\
    			345ADB   & 	[...]							&  [0.011,0.011,0.009,0.009,...] & 	   \\
    			12345AS  & 	[0.019,0.019,0.019,...]	&  [...] &  				    				   \\
    			...				&  		& &       	 	 	   	    
    		\end{tabular}
		\caption{Vectorization of Table~\ref{TblData}}
		\label{TblVect}
	\end{subtable}
	\bigskip
	\begin{subtable}{1\textwidth}
		\centering
    		\begin{tabular}{c|c|c|c}
    			User id     & G 								  & F & ...		 								  \\\hline
    			4421AB3  &  \cellcolor{gray}$(\alpha=0.93,\beta=10.99)$	& $(\alpha=1.06,\beta=0.25)$ &	 				\\
    			345ADB   & 	$(\alpha=0.69,\beta=6.95)$	&  $(\alpha=0.51,\beta=0.67)$ & 				 \\
    			12345AS  & 	$(\alpha=0.36,\beta=0.50)$	&  $(\alpha=1.64,\beta=0.89)$ &  				 \\
    			...				&  		& &       	 	 	   	    
    		\end{tabular}
		\caption{The valuation matrix representation of Table~\ref{TblVect} of after Beta estimations}
		\label{TblBeta}
	\end{subtable}
	\caption{Raw-data samples from the auction dataset ($Waterfall_A$) and their processed output.}
	\label{tab:dataPrep}
\end{table}

\end{document}